%% file: main.tex
\documentclass[10pt,twocolumn,letterpaper]{article}

\usepackage{cvpr}              

\usepackage{graphicx}
\usepackage{amsmath}
\usepackage{amssymb}
\usepackage{booktabs}

\usepackage{pythonhighlight}
\usepackage{multirow}
\usepackage{tipa}
\newlength{\Oldarrayrulewidth}
\newcommand{\Cline}[2]{%
  \noalign{\global\setlength{\Oldarrayrulewidth}{\arrayrulewidth}}%
  \noalign{\global\setlength{\arrayrulewidth}{#1}}\cline{#2}%
  \noalign{\global\setlength{\arrayrulewidth}{\Oldarrayrulewidth}}}

\usepackage[pagebackref,breaklinks,colorlinks]{hyperref}
\usepackage[capitalize]{cleveref}
\crefname{section}{Sec.}{Secs.}
\Crefname{section}{Section}{Sections}
\Crefname{table}{Table}{Tables}
\crefname{table}{Tab.}{Tabs.}

\def\cvprPaperID{11646} 
\def\confName{CVPR}
\def\confYear{2022}

\newcommand{\name}{PokeBNN\xspace}
\newcommand{\actfunc}{DPReLU\xspace}

\iftrue

\renewcommand{\ll}[1]{\textcolor{magenta}{\small [ll: #1]}}
\newcommand{\yz}[1]{\textcolor{brown}{\small [yz: #1]}}
\newcommand{\zz}[1]{\textcolor{blue}{\small [zz: #1]}}
\else

\renewcommand{\ll}[1]{\ignorespaces}
\newcommand{\yz}[1]{\ignorespaces}
\newcommand{\zz}[1]{\ignorespaces}
\fi

\newenvironment{tightlist}[4][\textbullet]
{
  \begin{list}{#1}
  {
    \setlength{\leftmargin}{#2}
    \setlength{\rightmargin}{#3}
    \setlength{\topsep}{0pt}
    \setlength{\parsep}{0pt}
    \setlength{\listparindent}{0pt}
    \setlength{\itemsep}{#4}
  }
}{
  \end{list}
}

\makeatletter
\newcommand{\printfnsymbol}[1]{%
  \textsuperscript{\@fnsymbol{#1}}%
}
\makeatother

\begin{document}

\newcommand*{\ARXIV}{}

\input{sections/title}
\input{sections/abstract}

\input{sections/intro}
\input{sections/related-work}
\input{sections/metric}
\input{sections/model}
\input{sections/experiment}
\input{sections/ablation}

\input{sections/conclusion}

\ifdefined\ARXIV
  {\small \bibliographystyle{ieee_fullname} \bibliography{egbib}}
  \input{sections/appendix}
\else
  {\small \bibliographystyle{ieee_fullname} \bibliography{egbib}}
\fi

\end{document}

%% file: sections/title.tex
\title{PokeBNN: A Binary Pursuit of Lightweight Accuracy}

\author{Yichi Zhang\thanks{Work performed while at Google, equal contribution.}\\
Cornell University\\
{\tt\small yz2499@cornell.edu}
\and
Zhiru Zhang\\
Cornell University\\
{\tt\small zhiruz@cornell.edu}
\and
Łukasz Lew\printfnsymbol{1}\\
Google Research\\
{\tt\small lew@google.com}
}
\maketitle

%% file: sections/abstract.tex
\begin{abstract}

Optimization of Top-1 ImageNet promotes enormous networks that may be  impractical in inference settings.
Binary neural networks (BNNs) have the potential to significantly lower the compute intensity but existing models suffer from low quality.
To overcome this deficiency, we propose PokeConv, a binary convolution block
which improves quality of BNNs by techniques such as adding multiple residual paths, 
and tuning the activation function.
We apply it to ResNet-50 and optimize ResNet's initial convolutional layer which is hard to binarize. We name the resulting network family  PokeBNN\footnote{Poke\textipa{/'p6kI/} is pronounced similarly to pocket. PokeConv, PokeBNN, and Pokemon are abbreviations of Pocket Convolution, Pocket Binary Neural Network, and Pocket Monster, respectively.}.
These techniques are chosen to yield favorable improvements in both top-1 accuracy and the network's cost.
In order to enable joint optimization of the cost together with accuracy, we define arithmetic computation effort (ACE), a hardware- and energy-inspired cost metric for quantized and binarized networks. 
We also identify a need to optimize an under-explored hyper-parameter controlling the binarization gradient approximation.

We establish a new, strong state-of-the-art (SOTA) on top-1 accuracy together with commonly-used CPU64 cost, ACE cost and network size metrics.
ReActNet-Adam~\cite{liu2021adambnn}, the previous SOTA in BNNs, achieved a 70.5\% top-1 accuracy with 7.9 ACE. 
A small variant of PokeBNN achieves 70.5\% top-1 with 2.6 ACE, more than 3x reduction in cost; 
a larger PokeBNN achieves 75.6\% top-1 with 7.8 ACE, more than 5\% improvement in accuracy without increasing the cost.
PokeBNN implementation in JAX/Flax~\cite{jax2018github, flax2020github} and reproduction instructions are open sourced.\footnote{
Source code and reproduction instructions are available in AQT repository: \href{https://github.com/google/aqt}{github.com/google/aqt}.
}

\end{abstract}

%% file: sections/intro.tex
\section{Introduction}
\label{sec:intro}

\textbf{A need for Pareto optimization.}
Deep learning research is largely driven by benchmarks and metrics. In the past a single metric per benchmark, \eg, top-1 accuracy on ImageNet, was sufficient. Today one needs to account for various model architectures, sizes, and computational costs. This promotes optimizing the Pareto frontier of a quality metric such as top-1 and another cost metric such as FLOPS, latency, or energy consumption. 

\begin{figure}
\vspace{-10pt}
    \centering
    \includegraphics[width=\linewidth]{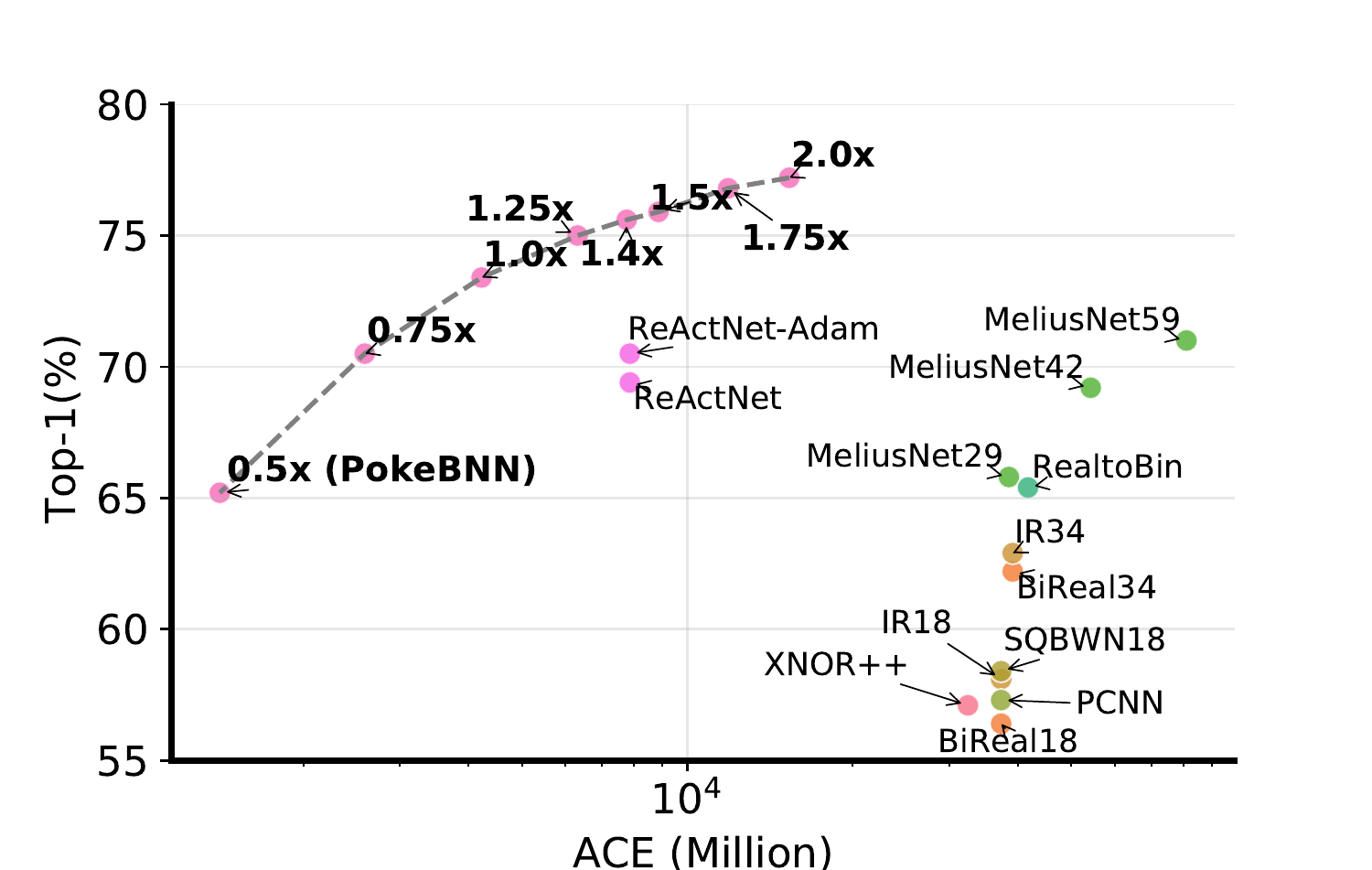}
    \caption{\textbf{Comparison of BNNs using top-1 and ACE (\cref{sec:ace}).}}
    \label{fig:energy-pareto}
    \vspace{-6pt}
\end{figure}

\textbf{The choice of the optimization metric.}
Since the industry is currently the main user of large-scale inference, the cost metric should be correlated to dollar cost per inference. 
As the ML hardware gets more mature, what becomes evident is that the energy use is the key metric proportional to the inference cost, especially in the data centers.
In \cref{sec:ace} we define a new proxy metric called \emph{arithmetic computation effort} (ACE), 
which aims to estimate inference cost abstracting of concrete ML hardware.

\textbf{The impact of quantization and binarization.}
In numerical formats such as int8 or float16, the less significant bits do not affect a network's output as much as the more significant bits. Yet, processing them consumes the same amount of energy. This might not be optimal.
The possible inference cost reduction (or equivalently performance improvement) with each halving of the quantization bits (\eg, 16b to 8b to 4b to 2b to 1b) is at least 2x (\eg, NVIDIA Ampere is 8x faster in int1 than in int8~\cite{nvidia2020ampere}) and up to 4x as estimated by the ACE metric. For comparison, this is significantly larger than the improvements yielded by upgrading the GPU or TPU by one or two generations. 

Binarization pushes this benefit to the extreme by replacing floating-point dot products with logical XNOR and bit counting operations.
If binary neural networks (BNNs) can reach high quality, they are likely to gain a large footprint for inference both in the data center and at the edge.

\textbf{BNN optimization is hard.}
Pioneering modern BNNs used to suffer from a more than 20\% top-1 accuracy gap compared to their floating-point counterparts~\cite{hubara2016bnn}. Only recently BNNs have become comparable in quality to the popular ResNet-18 model~\cite{Liu2020reactnet, liu2021adambnn}. 
One reason is that BNNs tend to have a chaotic, discontinuous loss landscape that renders their optimization challenging~\cite{li2017trainquantized, liu2021adambnn}.   
In fact, for the binarization to work one has to change many things compared to standard DNN practices. 
BNNs require multi-phase training,  
approximation of gradients, and various architectural adjustments that avoid binarization information bottlenecks.

\noindent \textbf{Our main contributions} are as follows: 
\begin{tightlist}{1em}{0.3em}{0.3em}
    \item We propose \textit{PokeConv}, a binary convolutional block that can substantially improve BNN accuracy. We replace most of the convolutions in ResNet~\cite{He2016resnet} with PokeConv.
    \item We propose \textit{PokeInit} block to replace ResNet's initial convolutional layer that is hard to binarize. PokeInit significantly reduces the network's cost. PokeInit and PokeConv form the foundation of the PokeBNN family.
    \item We optimize an under-explored clipping bound hyper-parameter in BNNs that controls the binarization gradient approximation. Ablation in \cref{sec:ablation} shows we gain more than 3\% in top-1 accuracy through this parameter. 
    \item We motivate and define a novel hardware and energy inspired cost metric called ACE, which is informed by inference costs on hardware yet at the same time it is agnostic to the existing hardware platforms. 
    ACE improves alignment of the research on energy-efficient neural networks and research on ML hardware.
    We use ACE to quantify the inference cost of PokeBNN.
    \item We empirically show that on ImageNet~\cite{olga2015imagenet} \name establishes the Pareto-SOTA of top-1 together with cost metrics: CPU64, ACE, and network size. We improve over the SOTA ReActNet-Adam by 5.1\% top-1 at the same ACE cost (\cref{fig:energy-pareto}).
\end{tightlist}

%% file: sections/related-work.tex
\section{Related Work}
\label{sec:related}

There is a large and active body of research investigating the training and acceleration of BNNs. We only review a subset of the past efforts that have a high influence on the network design presented in this paper. A comprehensive survey can be found in \cite{yuan2021comprehensive}.

\textbf{BNN feasibility.}
The pioneering works~\cite{hubara2016bnn, cheng2015bnntrain, kim2016bitwise} demonstrated the feasibility of BNNs.
They established the training framework for neural networks with binarized weights and activations and demonstrated promising results on small datasets such as MINIST and CIFAR-10. However, their preliminary ImageNet results show a large top-1 accuracy drop from 62.5\% to 36.1\% on AlexNet~\cite{alex2012alexnet} and from 68.9\% to 47.1\% on GoogleNet~\cite{szegedy2015googlenet}.

\textbf{Multi-phase training.}
A key effective technique is the multi-phase training~\cite{Bulat2019pose, Martinez2020realtobinary, Liu2020reactnet, liu2021adambnn}, where one starts with training an unquantized model and only later enables binarization. 
Some approaches employ a three-phase training — from the unquantized version, to binarized activations only, to binarized weights and activations~\cite{Martinez2020realtobinary}. Knowledge distillation is another technique that has commonly been used to improve the accuracy of BNNs~\cite{Bulat2019pose, Martinez2020realtobinary, Liu2020reactnet, liu2021adambnn}.

\textbf{BNN architecture.}
Another comprehensive line of work explores architectural changes to strive for better model quality. Many of them aim to incur negligible compute and parameter overhead. For example, a channelwise real-valued rescaling of the binarized tensors can effectively mitigate the quantization loss~\cite{rastegari2016xnornet, bulat2019xnornetplus, akhauri2019hadanets}. Connecting the unquantized input activations of a binarized convolutional layer to its output with a shortcut enhances the gradient flow and the model representation capacity~\cite{Liu2018bireal}. 
Squeeze-and-excitation (SE)~\cite{hu2018se} is another computationally cheap technique that promises quality improvement on small convolutional models including BNNs~\cite{Martinez2020realtobinary}. FracBNN~\cite{zhang2021fracbnn} includes additional BatchNorm Layers~\cite{ioffe2015batchnorm} in a BNN to speed up convergence. Authors in \cite{Bulat2019pose} first show that using a PReLU function~\cite{He2015rectifiers} after each convolutional layer improves binary model quality. Along this line, it is recently reported that introducing learnable biases into the PReLU function leads to extra improvements in model accuracy~\cite{Liu2020reactnet, liu2021adambnn}. With the evolution, current BNNs have finally exceeded 70\% top-1. 

%% file: sections/metric.tex
\section{Arithmetic Computation Effort}
\label{sec:ace}

In this section we motivate and define ACE, 
which is designed to reflect neural network inference cost on idealized ML hardware implemented with CMOS methodology.

\textbf{ACE metric definition.}
ACE is defined as follows: 
\begin{equation}
\operatorname{ACE} = \sum_{\substack{i \in I, j \in J}} n_{i, j} \cdot i \cdot j \end{equation}
where $n_{i, j}$ is the number of multiply-accumulate operations (MACs) between a $i$-bit number and $j$-bit number and can be automatically derived from model structure. $I$ and $J$ are sets of all bitwidths used in the inference of a given neural network, typically $I = J = \{1,2,4,8,16\}$.

\textbf{The energy use is highly correlated with the total cost of the computation.}
The inference could be happening in a data center or on edge devices and it can be served from CPUs, GPUs or TPUs. For edge devices, the battery usage is the main concern, which makes the energy use a key bottleneck in many ML applications. In the case of data centers, surprisingly, energy is also the main cost driver. 
In order to run inferences in a data center, one needs to pay for: hardware, electricity and power provisioning, and other infrastructure costs. 
A GPU card may cost 1000 USD and be used for 3-5 years consuming 400W. Electricity bill at 65\% utilization and 15 cents per kWh for three years would amount to 0.4kW * 24h * 365 * 3 * 0.65 * 0.15 USD/kWh = $\sim$1000 USD as well. Interestingly, the cost of the power provisioning in data centers (cooling, transformers, batteries, backup generators) is reported to be more than twice that of the electricity bill (at least in case of Google data centers)~\cite{jouppi2021lessons}. Also, a correlation of ML chip cost is reported to be over 90\% with its TDP. Overall, the cost of running inferences is indeed mainly driven by the energy consumption. 

\textbf{The bulk of the computation energy usage is in arithmetic operations energy.}
Contrary to classic CPUs, ML hardware running inference spends a high fraction of its energy on the actual arithmetic (e.g., multiplications, additions, other functions). For instance, in the case of TPUs, the cost of computation control is amortized over enormous SIMD sizes of 16K to 64K~\cite{jouppi2021lessons, jouppi2017tpus}. This is usually achieved using systolic arrays~\cite{kung1978systolic}. In stark contrast, CPUs have a typical SIMD size of 4 to 32 (e.g., SSE, AVX). 
We discuss other non-arithmetic energy sinks in the appendix in a full version of paper.

\textbf{Arithmetic operation energy is proportional to the number of active bit-adders.}
To multiply two unsigned integers $a < 2^I$, $b < 2^J$, one first computes a value of $I \cdot J$ bits using logical AND operations and sum them in groups:
\begin{equation}
\label{eq:bit-distribution}
\sum_{0\leq i < I}a_i 2^i\sum_{0\leq j < J}b_j 2^j =
\sum_{\substack{0\leq i < I \\ 0\leq j < J}} \left(a_i \land b_j\right) 2^{i+j} 
\end{equation}
In order to evaluate the sum, one uses bit-adders, carefully taking into account to add bit triplets within one significance group. 
Bit-adder sums three bits and outputs a two bit result: $p_1+p_2+p_3 = 2q_1 + q_2$ where $ p_i, q_i \in \{0, 1\}$. 
Bit-adders are the main building block of all  multipliers and adders. 
Each adder removes one bit from the pool, so taking into account addition into the accumulator (AC in MAC), a multiplication will activate $I \cdot J$ bit-adders.\footnote{While there are many orders in which one can construct adder trees (\eg,  Wallace tree~\cite{wallace1964multiplier}, Dadda tree~\cite{dadda1965parallelmultiplier}), affecting latency and clock speed, the particular order has a limited effect on the energy use.}
Notably, circuits that are not switching leak negligible amounts of energy, so one only pays for what they use.
One may verify that the number of active bit-adders is measured by ACE.

\textbf{CPU64 metric.}
Previous BNN research typically use $\text{FLOPs} + \frac{1}{64}\text{BOPs}$ as a cost metric~\cite{rastegari2016xnornet, Liu2018bireal, Liu2020reactnet, liu2021adambnn}. It was motivated by the fact that one 64-bit CPU register can do 64 BOPs in one cycle, compared to one float64 (double precision) operation per cycle. We extend CPU64 to int4 and int8 formats using coefficients 1/16 and 1/8, respectively.

\input{tables/tab-ace-new}

\textbf{Independent verification of energy use.}
Remarkably, the actual energy measurements on Google TPUs hardware are reasonably correlated with the ACE metric, grounding it in reality. \cref{tab:ace} reproduces energy measurement reported by Google and Horowitz~\cite{jouppi2021lessons, horowitz2014energy} on 45nm and 7nm process node and attaches both ACE and CPU64 metrics.
Interestingly, bfloat16 and to a large extent float16 and float32 are also well correlated with ACE both in 45nm and 7nm process nodes. We therefore choose to not special-case the ACE formula for MAC cost on floating-point formats.

\textbf{Implementation of high precision with binary arithmetic.}
If we interpret $a_i, b_i$ as binary matrices and $a_i \wedge b_i$ as binary matrix multiplication, then \cref{eq:bit-distribution} can be used to implement higher precision matrix multiplication on binary hardware. 
The cost of that emulation is $I\cdot J$, which is consistent with ACE metric. 
The result holds for all linear operations including convolution.

\textbf{Comparison to other metrics.}
Informed by the arithmetic energy use, ACE for MACs of N-bit and N-bit is quadratic in N as opposed to our CPU64 extension which is linear in N.
ACE generalizes FLOPS and CPU64 allowing for evaluation of mixed quantization models. 
ACE allows for evaluation of MACs with different bitwidths for weights and activations. This is useful as one of them is often much easier to quantize or binarize.
ACE is informed by CMOS hardware design and manufacture constraints yet at the same time is hardware target agnostic. 
With that we aim to better predict the performance of energy-efficient neural networks on the future ML hardware.
This is an advantage over popular methods of tuning the model for latency on GPUs or mobile hardware such as smartphones~\cite{sandler2018mobilenetv2, howard2019mobilenetv3}. 

%% file: tables/tab-ace-new.tex
\begin{table}[ht]
\footnotesize
  \centering
  \caption{\textbf{ADD/MUL energy use in femto-Joules (fJ)~\cite{jouppi2021lessons, horowitz2014energy}, and the corresponding CPU64 and ACE metrics.
  The correlation coefficient between ACE and the sum of ADD and MUL energy is 0.992 for 7nm and 0.946 for 45nm,
  whereas the CPU64-energy correlation is much smaller: 0.703 for 7nm and 0.724 for 45nm.
  }} 

  \vspace{-10pt}
  \label{tab:ace}
  \begin{tabular}{@{}l|rr|rr|rr@{}}
    \Cline{0.8pt}{1-7}
            & \multicolumn{2}{c|}{\textbf{ADD Energy (fJ)}}         & \multicolumn{2}{c|}{\textbf{MUL Energy (fJ)}}         & \multicolumn{2}{c}{\textbf{MAC}} \\ 
    \cline{2-7}
            & \textbf{45nm} & \textbf{7nm} & \textbf{45nm} & \textbf{7nm} & \textbf{CPU64} & \textbf{ACE}  \\
    \cline{1-7}
    float32 & 900     & 380    & 3700     & 1310 & 1 & 1024 \\
    \cline{1-7}
    float16 & 400     & 160    & 1100     & 340 & 1 & 256 \\
    \cline{1-7}
    bfloat16& -       & 110    & -     & 210 & - & 256 \\
    \cline{1-7}
    int32   & 100     & 30     & 3100     & 1480 & - & 1024  \\
    \cline{1-7}
    int8	& 30	     & 7	  & 200	  & 70 & 1/8 & 64   \\
    \cline{1-7}
    int4	& -	     & -	  & -	  & - & 1/16 & 16   \\
    \cline{1-7}
    int2    &	- & -	  & -	  & - & 1/32 & 4   \\
    \cline{1-7}
    binary  & -       & -	  & -	  & - & 1/64 & 1   \\
    \Cline{0.8pt}{1-7}
  \end{tabular}
\end{table}

%% file: sections/model.tex
\section{\name} \label{sec:model}

In this section, we introduce the design methodology of PokeBNN family. 
As a preliminary, we first define the quantization and binarization math used throughout the design.
We then introduce PokeConv --- a binarization friendly convolution replacement, and PokeInit --- a quantized and cost-optimized initial layer replacement.  Finally, we combine the proposed techniques and use ResNet as a template to present the entire PokeBNN architecture. We quantize the final layer to 8 bits, in effect, all the linear and convolutional layers are quantized to 8, 4 bits, or binarized.

\subsection{Quantization and Binarization Equations}
While both quantization and binarization methods are well studied~\cite{Abdolrashidi2021pareto, hubara2016bnn}, we summarize them for completeness.
In case of binarization, the clipping bound $B$ is usually hardcoded to 1~\cite{hubara2016bnn} or 1.3~\cite{Bethge2021meliusnet}. In \cref{sec:ablation} we show the importance of optimizing $B$. 

\textbf{Quantization.}
In order to use energy-efficient integer and binary convolutions and matrix multiplications, one needs to convert floating-point numbers into integers.
We define the casting operation as follows:
\begin{align*}
\operatorname{clip}(x, x_{min}, x_{max}) &= \operatorname{min}(x_{max}, \operatorname{max}(x_{min}, x)) \\
\operatorname{int}_b(x) &= \operatorname{round}(\operatorname{clip}(x, -C_b+\epsilon, C_b-\epsilon))
\end{align*}
where $b$ denotes the bitwidth, $C_b=2^{b-1}-0.5$ is the end point of the quantization grid, 
and $\epsilon$ is a small floating-point number making sure that the rounding avoids overflow.
For unsigned values, one uses 
$\operatorname{uint}_b(x)=\operatorname{floor}(\operatorname{clip}(x, 0, 2^b-\epsilon))$. For simplicity, we will focus on the signed case in the subsequent discussions.

During the backpropagation, the round function is ignored, \ie, $\frac{d\operatorname{round}}{dx}(x) = 1$.
This is known as the straight-through estimator (STE)~\cite{hubara2016bnn}. 
The derivative of clip operation is the usual 1 inside of the clip interval and 0 outside.

Applying casting directly to the arguments of convolution is inappropriate as their dynamic range can be different than the clipping bounds. Instead, one assumes (or estimates) bound $B$ based on the distribution of argument and then appropriately rescale it: 
\begin{equation}
\label{eq:quant-scheme}
    Q_b(x) = \operatorname{int}_b(x\cdot\frac{C_b}{B}) \cdot \frac{B}{C_b}
\end{equation}
One may note that the gradient is:
$\frac{dQ_b}{dx}(x) = \mathbf{1}_{x \in (-B, B)}$.

For non-binary activations, we obtain $B$ by calculating the maximum absolute value in a batch, and using exponentially moving average ($\alpha=0.9$). Importantly, we freeze $B$ when we enable activation quantization. Without freezing, we observe a feedback loop leading to inferior results or divergence. For binary activations, we show in \cref{sec:ablation} that the value of $B$ makes a remarkable impact on the model quality. We use a fixed $B=3$ in the experiments.

For all weights we use output-channel-wise bounds: $B_o = max_i \lvert w_{i,o} \rvert$ (where $i,o$ are indexing input and output channels). They are never frozen.
As equations indicate, we do not center the distribution. 
It is sufficient to just scale it.

\textbf{Binarization.}
When $b=2$, \cref{eq:quant-scheme} yields ternarization, but for binarization ($b=1$) it would round every number to 0. 
Instead we use $Q_1(x) = \operatorname{sign}(x)$. Its gradient is the same as that of \cref{eq:quant-scheme}.
Importantly, while the forward pass does not depend on bound $B$, the gradient does.
There is a line of work that studies a continous gradient estrimators to the discrete functions above~\cite{zhou2018dorefanet, gong2019dsq, qin2020irnet, Liu2018bireal}.

\subsection{PokeConv}
\label{sec:pokeconv}

\begin{figure}[h]
    \begin{minipage}{0.15\textwidth}
        \centering
        \includegraphics[width=0.9\linewidth]{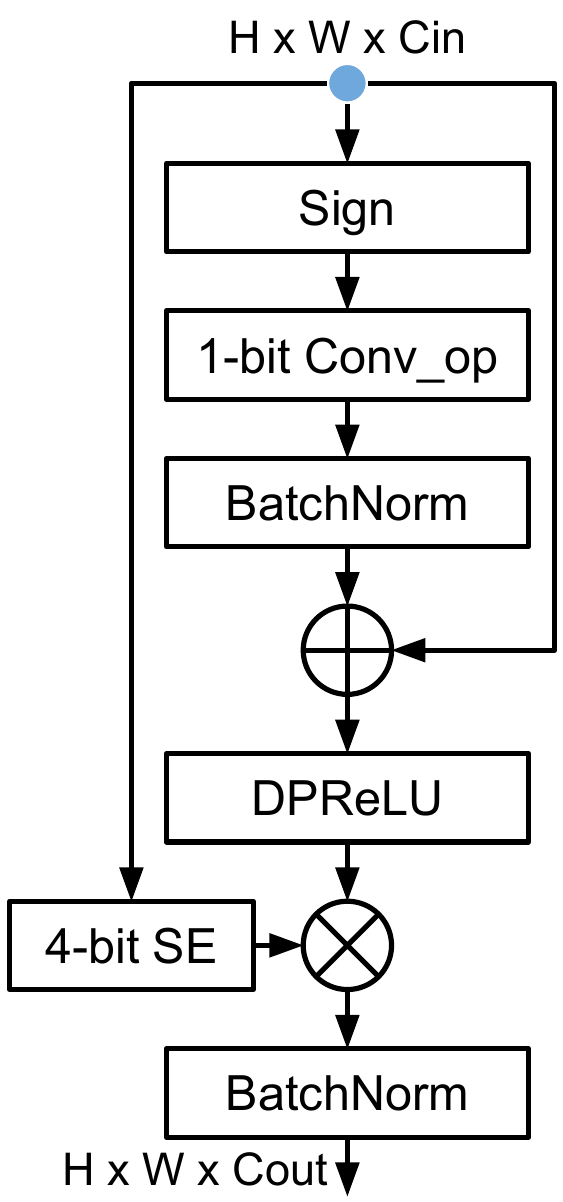}
    \end{minipage}
    \begin{minipage}{0.32\textwidth}
        \centering
        \begin{python}
def PokeConv(
    x, r1, conv_op, ch, st=1):
  r = x
  x = Conv_op(x, ch, st)
  x = BatchNorm(x)
  x = ReshapeAdd(x, r, pad_ch)
  # The next line has an effect
  # only in the 3rd PokeConv call
  x = ReshapeAdd(x, r1, tile_ch)
  x = DPReLU(x)
  x = x * SE_4b(r, ch)
  x = BatchNorm(x)
  return x
        \end{python}
    \end{minipage}
\caption{\textbf{PokeConv building block} --- "r1" is the original ResNet shortcut; "ch" is the number of channels or features; and "st" is stride.}
\label{fig:pokeconv}
\vspace{-6pt}
\end{figure}

We now propose the core convolutional (Conv) building block in our BNN, \textit{PokeConv}.
\cref{fig:pokeconv} shows its diagram and the corresponding pseudocode. The design of PokeConv strictly follows the goal of optimizing the accuracy and ACE trade-off. The additional operations around the binarized Conv layer are designed to be computationally lightweight and to help BNN training converge faster.

\textbf{Adding residuals around each binary convolution.}
We connect the unquantized input activations to the output of the Conv and BN combination with a shortcut.
Since the Conv layer is binarized, this shortcut is important as it removes the information bottleneck to the succeeding layer~\cite{Liu2018bireal}.

A practical question emerges naturally --- how do we design the shortcut if there is a mismatch between the input and output channel or spatial dimensions? Unfortunately, there is a lack of study on the general solution to it. A commonly-used method is adding a 1$\times$1 Conv layer with proper strides~\cite{He2016resnet}. However, this should be avoided since these layers are needed around most of the PokeConv blocks and they will increase ACE tremendously.
ReActNet~\cite{Liu2020reactnet} proposes to duplicate activations when the number of channels doubles, although this is only feasible when the channel number is exactly doubling.

We propose to use a simple zero padding when the number of channels expands.
Namely, given a channel expansion factor of $K$, an input tensor $x$ with $n$ channels is padded as follows:
$\operatorname{pad}(x)_i = x_i \cdot \mathbf{1}_{i < n}$.
We find that zero padding works the best for local shortcuts, and tiling proposed by ReActNet~\cite{Liu2020reactnet} works the best for the original ResNet shortcuts. Using a single method in both cases yields inferior results.

When the channel number decreases by a factor of $K$, we use an average pooling of the neighboring $K$ channels:
$\operatorname{avg\_ch}(x)_i = \frac{1}{K} \sum_{0\leq k < K} x_{i\cdot K + k}$.

On the spatial dimension, we use an average pooling on the shortcut for downsampling.
The pseudocode of aggregating the spatial and channel reshaping on residuals (the argument "r") is shown below:
\begin{python}
def ReshapeAdd(x, r, expand_ch_op):
  if r is None: return x
  if r.ch < x.ch: r = expand_ch_op(r, x.ch)
  if r.ch > x.ch: r = avg_ch(r, x.ch)
  if r.shape != x.shape: 
    r = avg_pool_3x3(r, x.ch, st=2)
  return x + r
\end{python}

\textbf{Using binarization-friendly nonlinearity.}
While the binarization function is nonlinear itself, inserting additional nonlinearity can further improve the BNN model quality as reported by prior studies~\cite{Bulat2019pose, Liu2020reactnet}, especially if the function learns to shift and reshape the activation distribution~\cite{Liu2020reactnet}.

We propose to add a nolinear function --- Dynamic PReLU (\actfunc)~\cite{ngoc2020dprelu} after the residual addition, defined as follows:
\begin{equation}
    \text{\actfunc}(x) := \left\{\begin{matrix} \eta(x-\alpha)-\beta & x-\alpha>0 \\ \gamma(x-\alpha)-\beta & \text{otherwise} \end{matrix}\right.
\label{eq:actfunc}
\end{equation}
Here $\alpha$, $\beta$, $\gamma$, and $\eta$ are all channelwise learnable parameters. They are initialized to 0, 0, 0.25, 1.0, respectively.
Aside from the activation shifting, \actfunc has learnable slopes on both linear pieces. It introduces an additional reshaping flexibility compared to RPReLU proposed by ReActNet~\cite{Liu2020reactnet}. 

\textbf{Squeeze-and-excitation (SE)~\cite{hu2018se} helps scaling.}
SE is a computationally cheap technique that improves model quality.
It allows the network to incorporate global knowledge on given inputs.
BNN such as real-to-binary~\cite{Martinez2020realtobinary} uses a different variant of SE as well. 

\begin{figure}[h]
    \begin{minipage}{0.21\textwidth}
        \centering
        \includegraphics[width=0.7\linewidth]{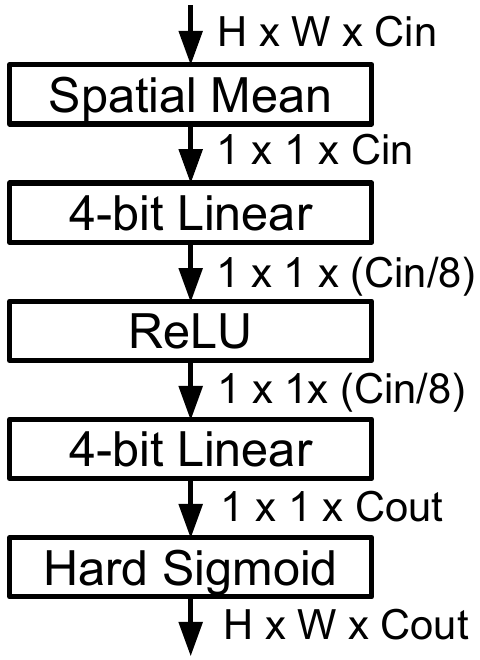}
    \end{minipage}
    \begin{minipage}{0.26\textwidth}
        \centering
        \begin{python}
def SE_4b(x, ch)
  s = SpatialMean(x)
  s = Linear_4b(s, x.ch/8)
  s = ReLU(s)
  s = Linear_4b(s, ch)
  s = ReLU6(s+3)/6
  return s
        \end{python}
    \end{minipage}
\caption{\textbf{Squeeze-and-Excitation: quantized to 4-bit.}}
\label{fig:SE}
\vspace{-8pt}
\end{figure}

In our design we apply an SE block as used in MobileNetV3~\cite{howard2019mobilenetv3}. 
The diagram and pseudocode are shown in \cref{fig:SE}.

Importantly, the ACE metric reports that SE dense layers incur a non-negligible cost.
We therefore propose to quantize the SE blocks to 4 bits, and set the hidden projection length to 1/8 of the input.
Experiments indicate that this modification incurs no accuracy loss.

\textbf{Additional normalization layers.}
We place an additional BatchNorm layer~\cite{ioffe2015batchnorm} on the output (\ie, before the residual split).
FracBNN~\cite{zhang2021fracbnn} suggests this modification in order to speed up the convergence.
This extra layer is even more important for \name. Firstly, its bias term learns to shift the distribution properly so that it balances the sign activations to the next PokeConv layer. Secondly, its adjustment on the distribution allows the aggressive 4-bit quantization of the first SE dense layer. Moreover, it normalizes the shortcut around binary convolutions and facilitates the gradient flow.

\textbf{Limitations.}
We use the default zero padding for Conv layers, which introduces a third value for a small fraction of pixels. We believe that most of the prior works suffer from the same limitation.
Padding with alternating 1 and -1 proposed and evaluated in FBNA~\cite{guo2018fbna} would resolve this limitation. 
\subsection{PokeInit and Projection Layer Optimization}
\label{sec:lightweight-layers}
After replacing regular Convs with PokeConvs, we find two other components in ResNet-50 that are ACE-costly:
(1) the input 7$\times$7 Conv layer; (2) the 1$\times$1 projection Conv layers for shortcuts in downsampling blocks.
These layers are conventionally excluded from binarization~\cite{liu2021sabnn}.

\textbf{Replacing 1$\times$1 projection Conv layers with ReshapeAdd.}
The 1x1 projection Conv layers would incur 360 million MACs.

We propose to completely remove these downsampling projection layers and replace them with the ReshapeAdd function defined for shortcuts. We use tiling instead of zero-padding for channel expansion, \ie, for an input tensor $x$ that has $n$ channels, $\operatorname{tile}(x)_i = x_{(i \operatorname{mod} n)}$.

\textbf{PokeInit.}
The unquantized input layer alone requires 118 million MACs.
The main sources of the large number of MACs are (1) the 7$\times$7 kernel size and (2) the large output spatial resolution 112$\times$112.

\begin{figure}[ht]
    \begin{minipage}{0.12\textwidth}
        \centering
        \includegraphics[width=0.9\linewidth]{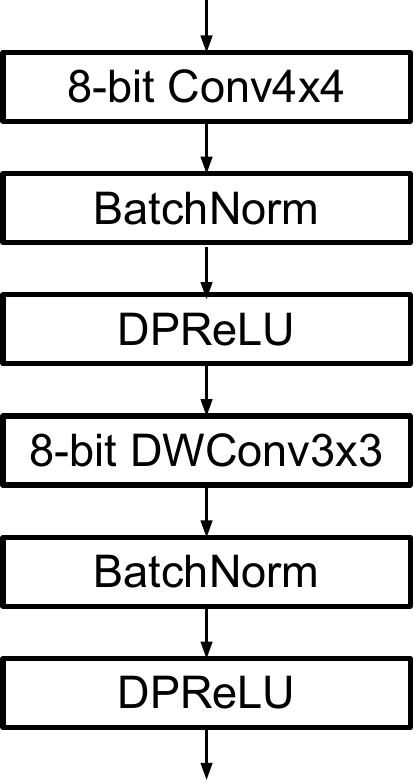}
    \end{minipage}
    \begin{minipage}{0.35\textwidth}
        \centering
        \begin{python}
def PokeInit(x):
  x = Conv4x4_8b(x, ch=32, st=4)
  x = BatchNorm(x)
  x = DPReLU(x)
  x = Conv3x3_depthwise_8b(x, ch=64)
  x = BatchNorm(x)
  x = DPReLU(x)
  return x
        \end{python}
    \end{minipage}
\caption{\textbf{PokeInit: quantized to 8-bit.}}
\label{fig:pokeinit}
\vspace{-8pt}
\end{figure}

To optimize the ACE cost, we fuse the stride-2 max-pooling with the first stride-2 convolution, yielding stride=4.
This reduces the output spatial resolution by 4$\times$. We then reduce the kernel size from 7$\times$7 to 4$\times$4. Further downsizing the kernel will lead to an information loss as there will be pixels not convolving with the kernels. To increase the receptive field of the input block, we follow it with a 3$\times$3 depthwise Conv layer.
We denote such an input layer combination as \textit{PokeInit}. 
Its pseudocode is shown in \cref{fig:pokeinit}. 

To further reduce the cost, we quantize PokeInit to 8 bits.
This optimization reduces the cost of the input layer from 118 million float MACs to 6.6 million int8 MACs. 

\subsection{Model Assembly}
We assemble PokeConv and PokeInit with ResNet-50 template with an 8-bit linear classifier head as shown in the pseudocode.

As discussed in the previous sections, we now apply PokeConv and PokeInit to the base ResNet-50, and remove the 1$\times$1 projection Conv layers therein.
Apart from those we also quantize the classifier to 8 bits.
A pseudocode of the final network is as follows:

\begin{python}
CH = 64 * M

def PokeBNN50(x):
  x = PokeInit(x)
  for i in range(16):
    st = 2 if i in (3, 7, 13) else 1 
    if i < 3: ch = CH
    elif: i < 7: ch = CH * 2 
    elif: i < 13: ch = CH * 4
    r = x
    x = PokeConv(x, None, Conv1x1_1b, ch, st=1)
    x = PokeConv(x, None, Conv3x3_1b, ch, st=st)
    x = PokeConv(x, r, Conv1x1_1b, 4*ch, st=1)
  x = SpatialMean(x)
  x = Linear(x)
  return x
\end{python}

Note that all Conv and linear layers do not use biases except for those that are not followed by BatchNorms. We follow ResNet~\cite{He2016resnet} to configure BatchNorm initializations.

%% file: sections/experiment.tex
\section{Experiments}
\label{sec:experiments}
In this section, we empirically evaluate \name on the ILSVRC12 ImageNet~\cite{olga2015imagenet} classification dataset with a resolution of 224$\times$224. We only apply random crop and flip as data augmentation. 

\subsection{Training Setup}

We conduct experiments on 64 TPU-v3 chips with a batch size of 8192. We use Adam optimizer~\cite{kingma2017adam} ($\beta_1 = 0.9, \beta_2 = 0.99$) with a linear learning rate decay. The initial learning rate is 6.4e-4. The weight decay is set to 5e-5 throughout the training. BatchNorm momentum is set to 0.9. To estimate the clipping bound $B$ for activation of non-binary quantized layers, we follow the method in \cite{Abdolrashidi2021pareto} and use exponentially moving average ($\alpha=0.9$) of maximum value in a batch.

We train \name for a total of 750 epochs and employ the two-phase training. 
We find that the first semi-unquantized phase is needed for only as little as 50  epochs.
4-bit and 8-bit activations are quantized at epoch 50.
All weights (8-bit, 4-bit, and binary) are quantized at epoch 50.
1-bit activations are always 1-bit during the training.

We use the knowledge distillation setting to train \name, which requires computing KL-divergence loss.
The modification is as simple as replacing the one-hot ground truth label with the teacher prediction. We use an 8-bit ResNet-50 as the teacher model. We also tried distilling from a vision transformer~\cite{dosovitskiy2021vit}, but surprisingly the result is similar.

In order to measure the accuracy, after decaying the learning rate to zero, we continue training for a few epochs. We observe both top-1 oscillating due to the batch normalization statistics being updated further. 
We find that training and evaluation top-1 are completely uncorrelated. Hence it would be unfair to follow a practice of taking the maximum top-1. All top-1 numbers in this paper are averaged top-1 over several epochs where learning rate is already zero. The difference between the mean and maximum accuracy is about 0.5\% to 1\%.

\subsection{Evaluation Results}

To have a fair comparison, we scale the number of output channels in a PokeConv block to change the model size (\eg, \name-2x means doubling the number of output channels). All results are collected in \cref{tab:big-table}. 
The standard deviation of the top-1 across 5 runs of PokeBNN-1x with different random weights is 0.034\%.

Importantly, for the prior work, we assume all FP32 operations could be replaced  by BF16 without accuracy loss. PokeBNN does not use FP32. Based on the data we analyze Pareto curve of accuracy vs. cost metrics, and compare the trade-off of \name to the baselines in the literature. We have several key observations thereof:

\textbf{{\name} establishes the SOTA Pareto frontier for BNNs under the ACE metric as visualized in \cref{fig:energy-pareto}.}
The accuracy of {\name} scales notably well with the model size. Though binarized from ResNet-50, scaling the number of channels by 2$\times$ in {\name} leads to a 77.2\% top-1 accuracy, slightly higher as the 4-bit ResNet-50 (\cref{tab:big-table}), yet with a more than 4.6$\times$ higher efficiency.

Most BNN models in the literature produce below 65\% top-1 accuracy on ImageNet. ReActNet~\cite{Liu2020reactnet} and ReActNet-Adam~\cite{liu2021adambnn} for the first time reach ResNet-18 level accuracy near 70\% by leveraging the MobileNet architecture~\cite{howard2017mobilenets}. 
With the same ACE budget as the current SOTA ReActNet-Adam, our \name-1.4x achieves 75.6\% top-1 accuracy, more than 5\% higher.
A small variant \name-0.75x has the same top-1 as ReActNet-Adam but reduces the ACE by more than 3$\times$.

Compared to the MeliusNet-59~\cite{Bethge2021meliusnet} variant that has the highest accuracy in the BNN literature (\cref{tab:big-table}), a large variant \name-2x is 6\% more accurate and meanwhile still 5.3$\times$ more efficient in ACE.

In addition, we test PokeConv on ResNet-18 architecture and observe that PokeBNN-0.5x Pareto-dominates it.

\begin{figure}[ht]
    \vspace{-15pt}
    \centering
    \includegraphics[width=\linewidth]{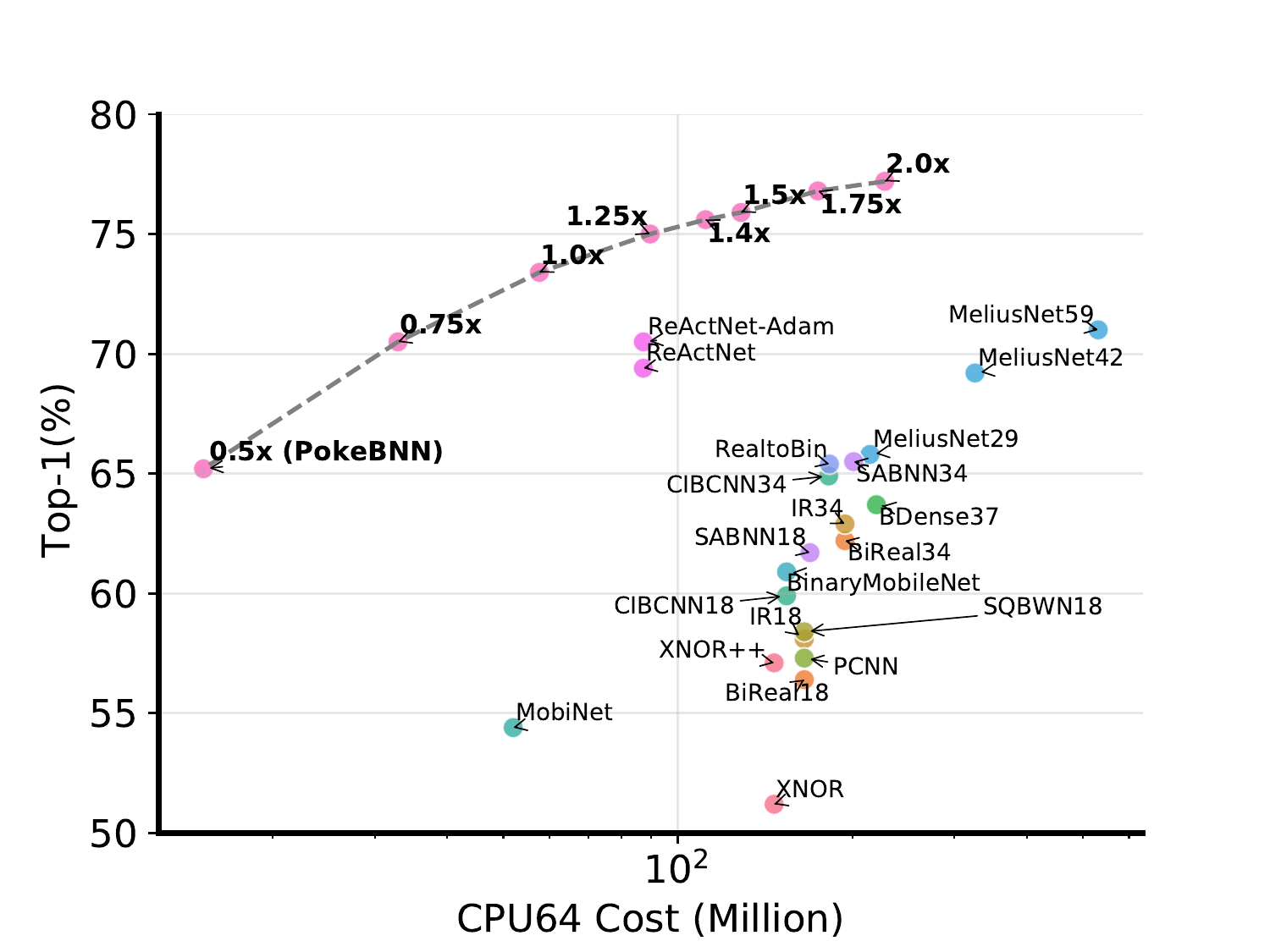}
    \caption{\textbf{Comparison of BNNs using top-1 and CPU64 cost.}}
    \label{fig:cpu64-pareto}
    \vspace{-10pt}
\end{figure}

\textbf{{\name} also establishes the SOTA Pareto frontier for BNNs under the commonly-used CPU64 metric.}  
We plot the Pareto curve using the widely adopted CPU64 metric in the literature. As shown in \cref{fig:cpu64-pareto}, the trade-off trend is roughly the same as compared to the proposed ACE metric. Notably, some BNNs (\eg, MeliusNet-29) show a less favourable trade-off under the CPU64 metric than the one in \cref{fig:energy-pareto}. This is because ACE captures the fact that a binary operation is more than 64$\times$ cheaper than a floating-point operation in terms of energy use. 

\begin{figure}[ht]
    \vspace{-10pt}
    \centering
    \includegraphics[width=\linewidth]{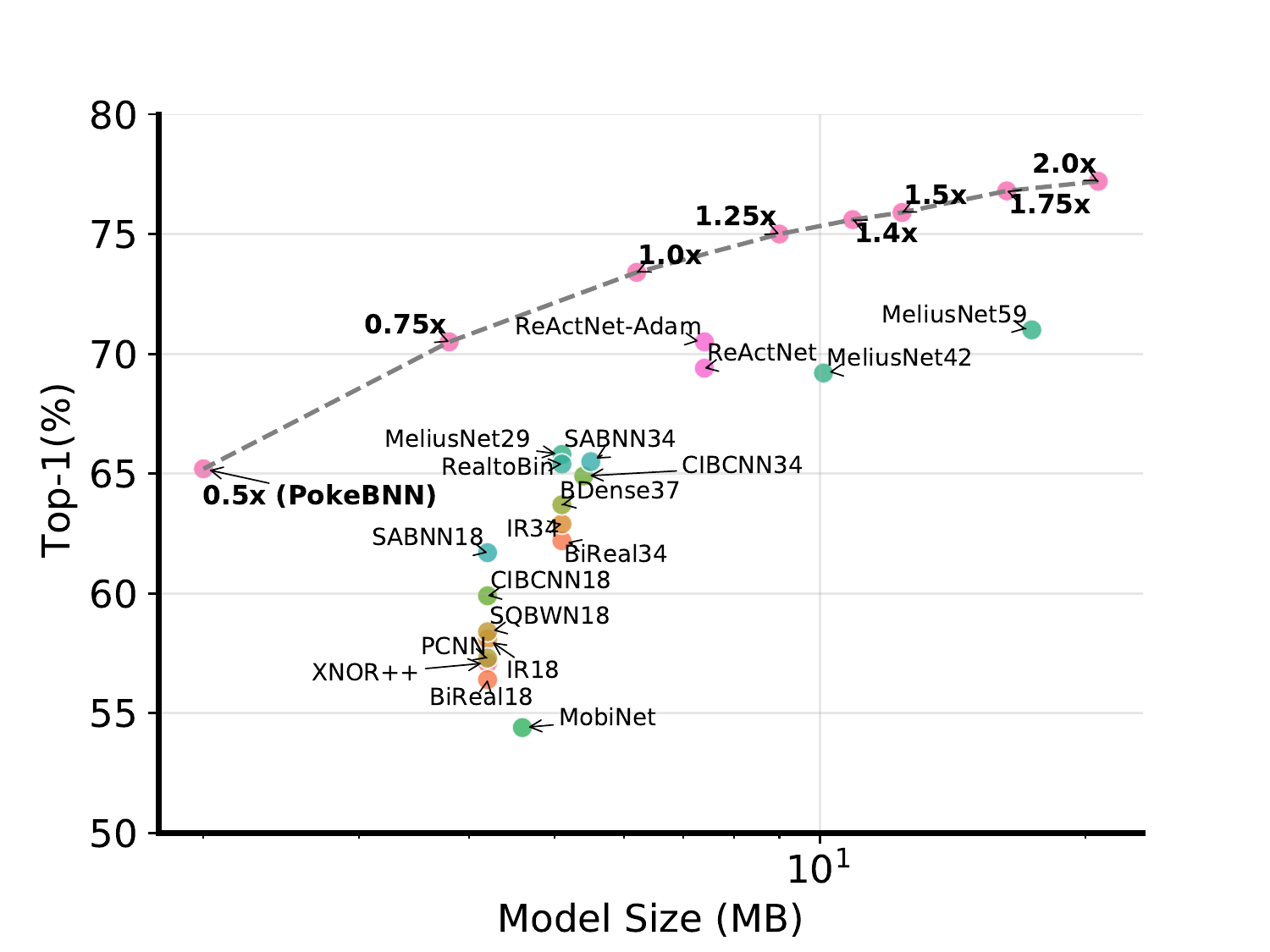}
    \caption{\textbf{Comparison of BNNs using top-1 and model size.}}
    \label{fig:memory-pareto}
    \vspace{-10pt}
\end{figure}

\textbf{{\name} outperforms prior BNNs on the size-accuracy trade-off.}
Model size is another important dimension that indicates the memory requirement. 
We therefore plot the model size vs. top-1 accuracy in \cref{fig:memory-pareto}, which shows that {\name} is also on the SOTA Pareto frontier when compared to the prior arts.

\input{tables/tab-big-table-new}

%% file: tables/tab-big-table-new.tex
\begin{table*}[ht]
\aboverulesep=0.1ex
\belowrulesep=0.1ex
 \footnotesize
  \def\arraystretch{0.5} 
  \addtolength{\tabcolsep}{-1pt}
  \centering
  \caption{\textbf{Final results and comparison to prior arts} --- When calculating ACE for FP32 operations, we assume they can be cast to BF16 without accuracy loss.
  ``--'' indicates unavailable data.
  The standard deviation of top-1 across 5 different seeds for \name-1.0x is 0.034\%.
  BF16 \name is a variant where all convolutions and dense layers are in BF16.
  The bottom four rows show the base models for context, all other models are binary.
  }
  \label{tab:big-table}
  \vspace{-10pt}
  \begin{tabular}{@{}lrrrrrrrrr@{}}
    \toprule
    \multirow{2}{*}{Model} & \multicolumn{5}{c}{MAC Operations ($10^6$)} & \multirow{2}{*}{ACE ($10^9$)} & \multirow{2}{*}{CPU64 ($10^6$)} & \multirow{2}{*}{Size (MB)} & \multirow{2}{*}{Top-1 (\%)} \\
    & FP32 & BF16 & INT8 & INT4 & Binary &  &  &  &  \\
    \midrule
    AlexNet-BNN~\cite{hubara2016bnn} & - & - & - & - & - & - & - & - & 36.1 \\
    \midrule
    GoogleNet-BNN~\cite{hubara2016bnn} & - & - & - & - & - & - & - & - & 47.1 \\
    \midrule
    XNOR-Net~\cite{rastegari2016xnornet} & 120 & - & - & - & 1700 & 32.4 & 146.6 & 4.2 & 51.2 \\
    \midrule
    XNOR-Net++~\cite{bulat2019xnornetplus} & 120 & - & - & - & 1700 & 32.4 & 146.6 & 4.2 & 57.1 \\
    \midrule
    Bi-RealNet-18~\cite{Liu2018bireal} & 139 & - & - & - & 1680 & 37.3 & 165.2 & 4.2 & 56.4 \\
    \midrule
    Bi-RealNet-34~\cite{Liu2018bireal} & 139 & - & - & - & 3530 & 39.1 & 194.2 & 5.1 & 62.2 \\
    \midrule
    IR-Net-18~\cite{qin2020irnet} & - & - & - & - & - & 37.3 & 165.2 & 4.2 & 58.1 \\
    \midrule
    IR-Net-34~\cite{qin2020irnet} & - & - & - & - & - & 39.1 & 194.2 & 5.1 & 62.9 \\
    \midrule
    SQ-BWN-18~\cite{dong2017SQBWN} & - & - & - & - & - & 37.3 & 165.2 & 4.2 & 58.4 \\
    \midrule
    PCNN~\cite{gu2019pcnn} & - & - & - & - & - & 37.3 & 165.2 & 4.2 & 57.3 \\
    \midrule
    BDenseNet37-Dilated~\cite{Bethge2019bdense} & - & - & - & - & - & - & 220.0 & 5.1 & 63.7 \\
    \midrule
    CI-BCNN-18~\cite{wang2019CIBCNN} & - & - & - & - & - & - & 154.0 & 4.2 & 59.9 \\
    \midrule
    CI-BCNN-34~\cite{wang2019CIBCNN} & - & - & - & - & - & - & 182.0 & 5.4 & 64.9 \\
    \midrule
    MobiNet~\cite{phan2020mobinet} & - & - & - & - & - & - & \textbf{52.0} & 4.6 & 54.4 \\
    \midrule
    BinaryMobileNet~\cite{phan2020binarymobile} & - & - & - & - & - & - & 154.0 & - & 60.9 \\
    \midrule
    MeliusNet-29~\cite{Bethge2021meliusnet} & 129 & - & - & - & 5470 & 38.5 & 214.5 & 5.1 & 65.8 \\
    \midrule
    MeliusNet-42~\cite{Bethge2021meliusnet} & 174 & - & - & - & 9690 & 54.2 & 325.4 & 10.1 & 69.2 \\
    \midrule
    MeliusNet-59~\cite{Bethge2021meliusnet} & 245 & - & - & - & 18300 & 81.0 & 530.9 & 17.4 & \textbf{71.0} \\
    \midrule
    Real-to-Binary Net~\cite{Martinez2020realtobinary} & 156.4 & - & - & - & 1676 & 41.7 & 182.6 & 5.1 & 65.4 \\
    \midrule
    SA-BNN-18~\cite{liu2021sabnn} & - & - & - & - & - & - & 169.0 & 4.2 & 61.7 \\
    \midrule
    SA-BNN-34~\cite{liu2021sabnn} & - & - & - & - & - & - & 201.0 & 5.5 & 65.5 \\
    \midrule
    SA-BNN-50~\cite{liu2021sabnn} & - & - & - & - & - & - & - & - & 68.7 \\
    \midrule
    QuickNetSmall~\cite{bannink2021larq} & - & - & - & - & - & - & - & \textbf{4.0} & 59.4 \\
    \midrule
    QuickNet~\cite{bannink2021larq} & - & - & - & - & - & - & - & 4.2 & 63.3 \\
    \midrule
    QuickNetLarge~\cite{bannink2021larq} & - & - & - & - & - & - & - & 5.4 & 66.9 \\
    \midrule
    ReActNet-A~\cite{Liu2020reactnet} & 11.9 & 0 & 0 & 0 & 4816.9 & \textbf{7.9} & 87.2 & 7.4 & 69.4 \\
    \midrule
    ReActNet-Adam~\cite{liu2021adambnn} & 11.9 & 0 & 0 & 0 & 4816.9 & \textbf{7.9} & 87.2 & 7.4 & 70.5 \\
    \midrule
    \textcolor{blue}{\name}-2.0x & 0 & 0 & 10.7 & 14.5 & 14412.2 & \textcolor{blue}{\textbf{15.3}} & 227.4 & 20.7 & \textcolor{blue}{\underline{\textbf{77.2}}} \\
    \midrule
    \textcolor{blue}{\name}-1.75x & 0 & 0 & 10.2 & 11.1 & 11037.1 & \textcolor{blue}{\textbf{11.9}} & 174.4 & 16.3 & \textcolor{blue}{\textbf{76.8}} \\
    \midrule
    \textcolor{blue}{\name}-1.5x & 0 & 0 & 9.7 & 8.2 & 8111.7 & \textcolor{blue}{\textbf{8.9}} & 128.5 & 12.4 & \textcolor{blue}{\textbf{75.9}} \\
    \midrule
    \textcolor{blue}{\name}-1.4x & 0 & 0 & 9.5 & 7.1 & 7037.2 & \textcolor{blue}{\textbf{7.8}} & 111.6 & 10.9 & \textcolor{blue}{\textbf{75.6}} \\
    \midrule
    \textcolor{blue}{\name}-1.25x & 0 & 0 & 9.2 & 5.7 & 5635.8 & \textcolor{blue}{\textbf{6.3}} & 89.6 & 9.0 & \textcolor{blue}{\textbf{75.0}} \\
    \midrule
    \textcolor{blue}{\name}-1.0x & 0 & 0 & 8.7 & 3.6 & 3609.5 & \textcolor{blue}{\textbf{4.2}} & 57.7 & 6.2 & \textcolor{blue}{\textbf{73.4}} \\
    \midrule
    \textcolor{blue}{\name}-0.75x & 0 & 0 & 8.2 & 2.0 & 2032.7 & \textcolor{blue}{\textbf{2.6}} & 32.9 & 3.8 & \textcolor{blue}{\textbf{70.5}} \\
    \midrule
    \textcolor{blue}{\name}-0.5x & 0 & 0 & 7.6 & 0.9 & 905.6 & \textcolor{blue}{\textbf{1.4}} & 15.2 & 2.0 & \textcolor{blue}{\textbf{65.2}} \\
    \Cline{1pt}{1-10}
    \midrule
    FP32 ResNet-50~\cite{He2016resnet} & 4089.2 & 0 & 0 & 0 & 0 & 1046.8 & 4089.2 & 97.3 & 76.7 \\
    \midrule
    BF16 ResNet-50~\cite{Abdolrashidi2021pareto} & 0 & 4089.2 & 0 & 0 & 0 & 1046.8 & 4089.2 & 48.6 & 76.7 \\
    \midrule
    INT4 ResNet-50~\cite{Abdolrashidi2021pareto} & 0 & 0 & 120.1 & 3969.1 & 0 & 71.2 & 263.1 & 13.1 & 77.1 \\
    \midrule
    BF16 \name & 0 & 3621.8 & 0 & 0 & 0 & 927.2 & 3621.8 & 50.3 & 79.2 \\
    \bottomrule
  \end{tabular}
  \vspace{-8pt}
\end{table*}

%% file: sections/ablation.tex
\section{Ablation Study}
\label{sec:ablation}

In this section we provide a detailed ablation on our proposed techniques. We measure the impact of each individual technique on \name-1.0x.

\textbf{Clipping bound ablation.}
The clipping bound $B$, as a hyperparameter, plays a major role in low-bitwidth quantization~\cite{Abdolrashidi2021pareto, choi2018pact}, but has rarely been explored in the past BNN research. In BNNs, although few works manually set the bound for binary activations $B=1.3$~\cite{Bethge2019bdense, Bethge2021meliusnet}, there is a lack of study on it and under most circumstances $B=1$ by default~\cite{hubara2016bnn, Liu2018bireal, Liu2020reactnet}.

In our experiments, we find that $B$ makes a remarkable impact on the BNN accuracy. We sweep $B$ for the binarized activations over a set of values ranging from 1.0 to 6.0. Each PokeConv has the same bound. Results are in \cref{tab:clip-bound}.

There is a 3.3\% accuracy gap between between $B=3$ and most commonly used value $B=1$.
We hypothesize that a larger clipping threshold improves Lipschitz constant of the loss surface and also reduces number of dead neurons (i.e., neurons with gradient zero).
This is consistent with the prior observation~\cite{Abdolrashidi2021pareto, choi2018pact} that the clipping bound is important for ultra low-bitwidth quantization.
\input{tables/tab-clip-bound}

\textbf{PokeConv ablation.}
We remove each component in PokeConv one at a time and study the impact.
The results are in \cref{tab:pokeconv-ablation}.
Removing \actfunc from PokeConv causes the largest accuracy drop, even larger than replacing PokeConv with the original binarized ResNet block. We hypothesize that this is because the change eliminates nonlinearity on both shortcuts, which impedes the model learning. We also replace \actfunc with RPReLU~\cite{Liu2020reactnet} and ReLU. This leads to a 0.2\% and 0.6\% top-1 degradation, respectively. Since 0.2\% is above 3 standard deviations (3 $\times$ 0.034\%), \actfunc indeed improves the model quality over the other two candidates. 

The other components (\ie, 4-bit SE, additional shortcuts and BatchNorms) all have at least 3\% top-1 impact on the model. Given they incur negligible overhead in ACE, they are favourable design choices.
\input{tables/tab-pokeconv-ablation}

We also experiment with adding back the 1$\times$1 projection Conv layers. The top-1 result is 73.5\%, only 0.1\% higher. Completely removing these layers is therefore sensible.

\textbf{PokeInit ablation.}
In \name-1.0x, we replace PokeInit with ResNet's original 7$\times$7 input Conv layer followed by a maxpooling. The top-1 is 73.5\%, only 0.1\% higher.
Given that PokeInit reduces the number of operations in the input layer by 18$\times$, it is a favourable trade-off.

We also experiment with removing the 3$\times$3 depthwise Conv layer in PokeInit. This results in a 73.1\% top-1. The depthwise layer trades 2.7\% of the total ACE cost for 0.3\% accuracy, which is also a fair trade-off.

\textbf{Precision ablation.}
Increasing the weight or activation precision in PokeConv from 1-bit to 4-bit results in a 75.2\% and 76.8\% top-1, respectively. Both of these variants have an ACE cost of 15, and both are significantly better than INT4 ResNet~\cite{Abdolrashidi2021pareto} but worse than \name-1.75x. This result indicates that binarization indeed allocates energy better than int4 formats.

%% file: tables/tab-clip-bound.tex
\begin{table}[ht]
\footnotesize
\addtolength{\tabcolsep}{-1pt}
  \centering
  \caption{\textbf{Impact of the activation clipping bound $B$ in the binarization function.}}
  \vspace{-8pt}
  \label{tab:clip-bound}
  \begin{tabular}{@{}cccccccc@{}}
    \toprule
    Clipping Bound $B$ & 1.0 & 1.3 & 2.0 & 3.0 & 4.0 & 5.0 & 6.0 \\
    \midrule
    Top-1 (\%) & 70.1 & 71.4 & 72.9 & \textbf{73.4} & 73.3 & 72.8 & 72.4 \\
    \bottomrule
  \end{tabular}
  \vspace{-4pt}
\end{table}

%% file: tables/tab-pokeconv-ablation.tex
\begin{table}[ht]
\footnotesize
\vspace{-4pt}
  \centering
  \caption{\textbf{Ablate each component in PokeConv.} "All" indicates replacing PokeConv with the original 1-bit ResNet Conv block.}
  \vspace{-6pt}
  \label{tab:pokeconv-ablation}
  \begin{tabular}{@{}lccccc@{}}
    \toprule
    Remove Module & SE & DPReLU & Shortcuts & BN & All \\
    \midrule
    Top-1 (\%) & 70.6 & 60.4 & 68.1 & 70.2 & 61.9 \\  
    \bottomrule
  \end{tabular}
  \vspace{-8pt}
\end{table}

%% file: sections/conclusion.tex
\vspace*{-0.2cm}
\section{Conclusion}

The main ingredients of PokeBNN: PokeConv, PokeInit, and the clipping bound ($B=3$), together establish a strong SOTA in the domain of cost-efficient networks.
ACE metric improves alignment of research on cost-efficient neural networks with future ML hardware.
Our results indicate that binarization may indeed be a good choice in cost-accuracy trade-off. 
The main price of these benefits is a 750-epoch long training.

\textbf{There are several unanswered questions.} 
How to take energy of memory access into account in a synthetic metric? 
How could the Poke architecture be further simplified or improved?
Could architecture templates different than ResNet-50 or perhaps neural architecture search yield significantly better networks? 

\textbf{Acknowledgements.} 
The authors would like to thank Catalyst, JAX, and Flax teams for valuable implementation, discussions, and suggestions on  \href{https://github.com/google/aqt}{AQT library}.
This work is supported in part by NSF Award \#2007832.

%% file: sections/appendix.tex
\section{Appendix}

\subsection{Cost of Elementwise Operations}
\label{sec:appendix-cost}

\begin{table*}[ht]
  \centering
  \begin{tabular}{@{}lrr@{}}
    \toprule
    \textbf{Layer Type} & \textbf{ADDs ($\times 10^6$)} & \textbf{MULs ($\times 10^6$)} \\
    \toprule
    \texttt{BatchNorm} addition & 17.9  & 17.9\\
    \midrule
    \texttt{DPReLU} & $3 \times 9.1$  & 9.1\\
    \midrule
    \texttt{ReshapeAdd:} \verb'avg_ch' & 5.4 & 1.7 \\
    \midrule
    \texttt{ReshapeAdd:} \verb'avg_pool_3x3' & 7.9 & 0.87 \\
    \midrule
    \texttt{ReshapeAdd:} residual addition (local + block) & 8.8 + 5.5 & 0\\
    \midrule
    \texttt{SE\_4b: SpatialMean} & 8.9 & $<0.01$ \\
    \midrule
    \texttt{SE\_4b:} activation functions & $<0.01$ & $<0.01$\\
    \midrule
    \texttt{SE\_4b:} final multiplication & 0 & 8.8 \\
    \midrule
    Global pooling before classifier & 0.1 & $<0.01$ \\
    \midrule
    \midrule
    Sum & 81.9 & 38.4 \\
    \bottomrule
  \end{tabular}
  \caption*{\textbf{Number of elementwise operations from unquantized layers.}}
  \label{tab:sops}
\end{table*}

It is usually assumed that elementwise operations such as residual connections, BatchNorm layers, or DPReLU activation functions have negligible computation compared to Conv or linear layers. 
These elementwise layers are therefore excluded when calculating the compute cost with CPU64 or ACE. 
BNNs in the literature also adopted the same assumption~\cite{Liu2018bireal, Liu2020reactnet, liu2021adambnn}.

However, the elementwise operations gradually become the cost bottleneck after the expensive Conv and linear layers are either binarized or quantized to ultra-low precision. 
Using PokeBNN-1.0x as a case study, we show the breakdown of the operation count in the table below. 

There are 81.9 million elementwise additions and 38.4 million elementwise multiplications in PokeBNN-1.0x. These operations remain unquantized and are executed in bfloat16. 
If there is no optimization, these operations will have $30.8\times10^9$ ACE and will completely dominate the cost.
Prior BNN works also suffer from the same issue~\cite{Liu2018bireal, Liu2020reactnet, liu2021adambnn}.
Described as follows, there are several ways in which we can alleviate the problem.

\textbf{Use integer arithmetic for additions.}
Most of the elementwise operations are additions. 
According to the energy numbers in Table 1 in the paper, additions in float16 are about 4 to 5 more expensive than additions in int32, and 13 to 23 times more expensive than int8. 
It would be challenging to represent all the ops in int8 without an automatic calibration algorithm,
but int16 should have wide enough dynamic range.

Assume that no activation needs an absolute value bigger than $2^7$, add the following operation can be applied on all the operations in the network:
$$Q_{16}(x) = \operatorname{clip}(\operatorname{round}(x\cdot2^8)\cdot2^{-8}, -(2^7 -1),  (2^7 -1))$$
This implements a low-cost fixed-point arithmetic that has int16 additions. 
Only 16 adders are needed so the ACE metric is 16 for each addition.

\textbf{Use narrow integer arithmetic for multiplication.}
High precision integer multiplication is costly.
All of the multiplications in PokeBNN inference are fixed-point.
The multiplication constants in \verb'avg_ch' are 0.5 or 0.25, these can be implemented by shifts with negligible energy cost.
In \verb'avg_pool_3x3' the constant is $\frac{1}{9}$, but we could approximate it by $\frac{1}{8}$ and use a shift as well.
It is the multiplications in BatchNorm, DPReLU and the final multiplication in \verb'SE_4b' that represent the biggest challenge.
In this appendix we assume that we assume that their 8-bit fixed-point representation of the fixed multipliers is accurate enough. The cost of int16 multiplied by int8 is ACE=$16\cdot8 = 128$.

\textbf{Operation fusion.}
All the unquantized operations in our network are linear as they consist of multiplications and additions. The one exception is the slope selection in DPReLU.
Nevertheless the size of the channelwise learnable parameters in DPReLU and BatchNorm is small. It is only a vector of length of the number of channels as they are shared between the spatial pixels. 
The same holds for the output of the SE layer.
One can check that the consecutive operations of DPReLU and SE output multiplications can be always fused with the neighboring BatchNorm as all of them are affine functions. Only a separate addition for slope selection needs to be preserved. Both scaling multiplication in the quantization operators can be fused as well. 
The cost of fused operations can be ignored for the inference. 
In the case of fusion of SE final multiplication scaling, computation of the coefficients for final affine transformation has to be dynamic.
Nevertheless, the cost of it is also negligible as it is shared between pixels and proportional only to the number of channels.

\textbf{Smaller Pooling Kernel.} Current average pooling layers for spatial downsampling uses 3$\times$3 kernels. 
One could also replace \verb'avg_pool_3x3' with \verb'avg_pool_2x2', a transformation similar to the first layer in PokeInit. It reduces the cost by 55\%.

\textbf{ACE estimation for pointwise operations.}
While the additions can be fused as described above, even without that the cost of them will be relatively low $ACE \leq 81.9 \cdot 10^6 \cdot 16 \approx 1.3 \cdot 10^9 $. 

Importantly, all the multiplications not in BatchNorms can be either replaced by shifting or fused into BatchNorm. The cost of the MULs in BatchNorm is $ACE = 17.9 \cdot 10^6 \cdot 128 = 2.3\cdot 10^9 $.

Overall the cost of elementwise operations is estimated to be $ACE = 3.6\cdot 10^9$. While this is less than the cost of convolutions and the final classifier in PokeBNN-1.0x ($ACE=4.2\cdot 10^9$), it is not negligible and should be taken into account in future research.

\subsection{ACE limitations}
\label{sec:appendix-ace-limit}

\textbf{Memory reads and writes are potentially major energy sinks.}
The energy cost of writing or reading data to DRAM is 50-150 higher than to SRAM, but in case of the inference both model and activations usually can (or have to) fit in SRAM.
We also note that if systolic array matrix multiplication circuit is big enough, for an inference, all inputs to the matmuls and convolutions have to be read exactly once. 
ACE does not estimate the cost of SRAM reads and writes into account. 
Also, as discussed in~\cite{jouppi2021lessons} the SRAM energy cost is not reducing as fast as the arithmetic cost between 45 nm to 7 nm chip manufacture process, thus making it incompatible with the goals of ACE.

\textbf{Data movement are potentially major energy sinks.}
Energy-intensive chips are limited to 2D due to a need of high area-to-volume ratio for the sake of power delivery and heat removal. Inability to co-locate various circuits and memories may force existence of long wires, so called buses. This problem is not fundamental as tiled chip designs are being explored and well suited for 2D (also 3D and higher) image processing as each tile may be responsible for part of image and communicate only with its neighbors. Also energy use of chip's clock-tree can be thought of a variant of the same problem. Taking such design considerations into account is far beyond the scope of ACE.

\textbf{ACE does not take into account "vector" operations.}
Activations, multiplication in BN, residual additions, bound scaling and clipping. All these operations are not taken into account neither by our application of ACE nor by most of the papers using CPU64.
The analysis on the these operations is described in Appendix: Cost of Elementwise Operations.

\textbf{Memory layout and padding.}
DNN operations that reshape or pad large patches of data are not modeled for energy. Networks like ShuffleNet[cite] might not be modeled well enough by ACE.

\textbf{Analog and in-memory computing.}
It is unclear to the authors how to model hardware and networks that perform arithmetic operations using analog, in-memory computing circuits.